# Gradient Descent Quantizes ReLU Network Features

Hartmut Maennel    Olivier Bousquet    Sylvain Gelly

Google Brain

**Abstract**

Deep neural networks are often trained in the over-parametrized regime (i.e. with far more parameters than training examples), and understanding why the training converges to solutions that generalize remains an open problem Zhang et al. [2017].

Several studies have highlighted the fact that the training procedure, i.e. mini-batch Stochastic Gradient Descent (SGD) leads to solutions that have specific properties in the loss landscape. However, even with plain Gradient Descent (GD) the solutions found in the over-parametrized regime are pretty good and this phenomenon is poorly understood.

We propose an analysis of this behavior for feedforward networks with a ReLU activation function under the assumption of small initialization and learning rate and uncover a quantization effect: The weight vectors tend to concentrate at a small number of directions determined by the input data. As a consequence, we show that for given input data there are only finitely many, "simple" functions that can be obtained, independent of the network size. This puts these functions in analogy to linear interpolations (for given input data there are finitely many triangulations, which each determine a function by linear interpolation). We ask whether this analogy extends to the generalization properties - while the usual distribution-independent generalization property does not hold, it could be that for e.g. smooth functions with bounded second derivative an approximation property holds which could "explain" generalization of networks (of unbounded size) to unseen inputs.

## 1 Introduction

Understanding the fundamental reasons why Deep Neural Networks (DNNs) generalize well even in the absence of explicit regularization has attracted a lot of interest Zhang et al. [2017], Wu et al. [2017], Dinh et al. [2017], Hardt et al. [2016], Keskar et al. [2016], Smith and Le [2018]. Common training practice include using Stochastic Gradient Descent (SGD), very large and deep networks, residual connections, ReLU activation functions, small weight initialization, ... but the role played by each of these components remains poorly understood.

Many works are focused on the generalization gap between SGD and GD, or small batch vs large batch Wu et al. [2017], Hardt et al. [2016], Keskar et al. [2016]. As noted by Wu et al. [2017], while there is indeed empirical evidence that SGD leads to networks that generalize better than GD, the difference is often within a few % accuracy. However, DNNs can easily fit arbitrary labels, leading to networks that can have arbitrary bad generalization properties while still perfectly fitting the training data Zhang et al. [2017], yet they generalize well under normal conditions. It is possible that GD combined with specific weight initialization (in particular small norm) explain a large part of the generalization properties of DNNs, and we explore this particular question.

We formally study the behavior of GD on one hidden layer ReLU neural networks, in the limit of infinitesimal initial weights and learning rate. Under this idealized model, we show that the dynamical system corresponding to the training procedure can be interpreted as a two steps process:

i) Without changing the loss significantly, the neuron weight vectors align to a *discrete* set of possible directions, which only depend on the training data, not on the size of the network.

ii) Loss is reduced.

This also implies the equivalence of the resulting network with a simpler network obtained by greedily adding one neuron at a time. There are a only a finite number of such "replacement networks", which are defined purely in terms of the training data.

For example, if we have $K$ training inputs that lie on a line, the resulting finitely many functions of the possible replacement networks are piecewise linear with at most $2K + 1$ kinks. In numerical simulations we often get just the linear interpolation or something close to it.

Also for higher dimensional inputs the resulting situation has similarity to linear interpolation, where choosing one of finitely many possible triangulations defines an interpolating function. This may also be true of the generalization properties: While neither linear interpolation nor these network functions satisfy the usual distribution-independent generalization property, we ask whether they have an approximation property when restricted to e.g. smooth functions with bounded second derivative.

(Note this would not be the usual known approximation property that neural networks *can* approximate any continuous function, but that under some condition Gradient Descent gives a result that *does* approximate the true function within specified error bounds.)

We then experimentally show that this idealized model partially corresponds to the training behavior of neural networks on toy data and MNIST.

**Related work:** In Brutzkus et al. [2017], a similar setup is studied but with leaky ReLU activiations (i.e. $\sigma(x) = \max(\alpha x, x)$ for $\alpha > 0$ instead of $\alpha = 0$ which we study here). The authors show that SGD can converge to a global optimum in a number of steps that is independent of the size of the network. Here we do not study the convergence speed, and we consider GD. We are more focused on the qualitative nature of the solution found by GD with a specific low norm initialization.

Other authors have discussed the appearance of two different phases in (stochastic) gradient descent training of neural networks. For example Shwartz-Ziv and Tishby [2017] show that under certain conditions, there is a first phase where the network is fitting the data and a second phase where the internal representation gets compressed (i.e. the information contained in the hidden layers about the input gets reduced). The quantization phenomenon that we highlight here can be thought of as a kind of compression of information. However, the surprising result is that this compression happens in the first phase, i.e. before the network starts to actually fit the data.

Note that this discussion doesn't take into account the norm of the weights nor the norm of the data points which are typically used to derive size-independent complexity measures of neural networks as in for example Neyshabur et al. [2014],Neyshabur et al. [2015],Golowich et al. [2018]. It would be interesting to explore whether one can refine such measures using the quantization effect discussed here.

## 2 General set up

We are looking at neural networks with ReLU activation function $\rho(t) := \max(0, t)$ that model functions $f$ that take input variables $\vec{x} \in \mathbb{R}^d$ and produce one output value $y \in \mathbb{R}$.

We learn with gradient descent from training examples $(\vec{x}_k, y_k)$, using the loss function
$$L(f) := \sum_{k=1}^{K} \ell(f(\vec{x}_k), y_k)$$

where $\ell(\hat{y}, y)$ is differentiable; we are mostly interested in either the square distance loss for regression $\ell(\hat{y}, y) := \frac{1}{2}(\hat{y} - y)^2$ or the cross entropy loss for classification, i.e. $y \in \{\pm 1\}$ and $\ell(\hat{y}, y) := \log\left(1 + e^{-y \cdot \hat{y}}\right)$.

For this paper, we restrict ourselves to one hidden layer:
$$f(\vec{x}) = b' + \sum_i w'_i \cdot \rho(b_i + \sum_j w_{ij} x_j)$$

We will use the generalization to a neural network with one more input, but no biases. We can express this with with an input vector $\vec{x} \in \mathbb{R}^d$ as

$$f(\vec{x}) = \sum_{i=1}^{n} w'_i \cdot \rho(\langle \vec{w}_i, \vec{x} \rangle) \tag{1}$$

We recover the special case above by setting the additional input to 1: If $d$ is the index of this fixed additional input, then the $b_i$ above becomes $w_{i,d}$. If we choose one of the weights, e.g. the last $\vec{w}_n$ such that $w_{n,j} = 0$ for all $j \neq d$, then the corresponding contribution is

$$w'_n \cdot \rho(\langle \vec{w}_n, \vec{x} \rangle) = w'_n \cdot w_{n,d} \qquad \text{if } w_{n,d} \geq 0,$$

so by choosing $w'_i$ and $w_{n,d} \geq 0$ such that $b' = w'_n \cdot w_{n,d}$, we also recover the above overall bias $b'$.

As our optimization algorithm we use the plain Gradient Descent. This means we initialize the $\vec{w}_i$ and $w'_i$, and then in each step we read in all the data $(\vec{x}_k, y_k)$ and update the weights with a learning rate $\epsilon$:

$$w'_i \to w'_i - \epsilon \cdot \frac{\partial}{\partial w'_i} L(f) \qquad \vec{w}_i \to \vec{w}_i - \epsilon \cdot \nabla_{\vec{w}_i} L(f)$$

We do not use early stopping, i.e. the trained network has the weights $\lim_{i \to \infty} w'_i$ and $\lim_{i \to \infty} \vec{w}_i$.

## 3 First empirical observations

### 3.1 The network often produces a simple function

We train a large network (e.g. 200 hidden neurons) to interpolate five points $x = -2, -1, 0, 1, 2$ with values $y = 1, 0, 1, 0, 1$. Due to the large amount of freedom the network has, we may expect the interpolation to look like the left graph, but in fact what we get "usually" is the simple linear interpolation on the right hand side:

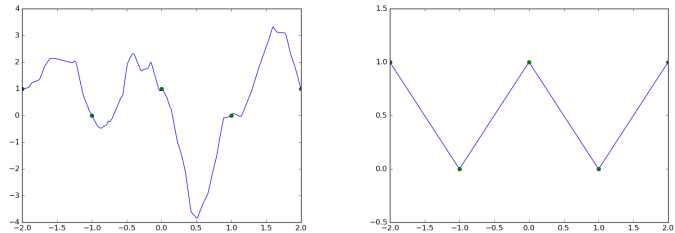

In fact, we also get the same picture for deep networks, but in this paper we only analyze the case of one layer.

We see a similar behaviour in two dimensions: We give the corners of a square to a network to approximate the values (clockwise) 0,1,2,3. We would expect something like the left graph, but we get a simple function that could be implemented with 2 hidden neurons (right graph).

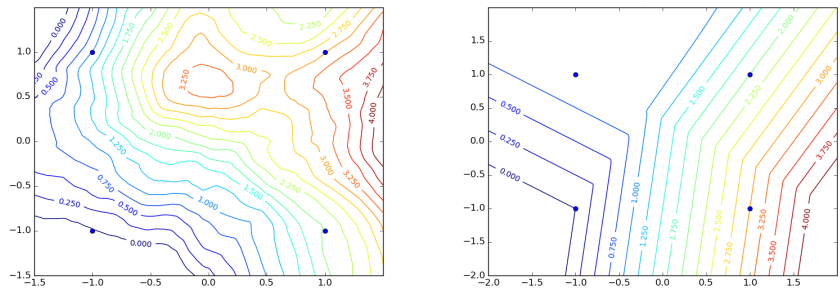

To explain this observation mathematically, we first have to analyze under which conditions it occurs.

## 3.2 Learning rate and initialization

Without restriction to learning rate $\epsilon$ and initialization we cannot say much about the interpolating function to which the network converges - in fact the left hand graphs above were obtained with the same network as the right hand graphs, but with larger initial values for the weights.

For the learning rate there is an obvious candidate for predictable behaviours: We look at the case of infinitesimal learning rate $\epsilon \to 0$ - this converts the dynamical system with discrete time steps into a continuous system which can be described by differential equations.

To understand the dependence on the initialization we fix some $\vec{v}_i \in \mathbb{R}^d$ and $v_i \in \mathbb{R}$ (e.g. sampled randomly from some normal distribution with some fixed seed of our random number generator) and initialize the weights as $\vec{w}_i := \lambda \cdot \vec{v}_i$, $w'_i := \lambda \cdot v'_i$ with different $\lambda$ (or equivalently, for our randomly sampled weights we adjust the standard deviation, but keep the same seed).

When we apply this rescaling with $\lambda \to 0$, we get as functions after convergence:

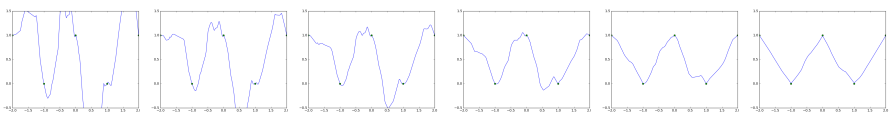

(Here $\lambda = 0.8, 0.6, 0.5, 0.4, 0.3, 0.2$, for $\lambda \leq 0.1$ the pieces look linear like before) So to get a behaviour which we can analyze mathematically, and which is still not too far from what we may do in practice, we will consider the case of $\lambda \to 0$.

## 3.3 Kinks move first and align

Another empirical observation in 1d is that the ReLU kinks move while the training loss stays approximately constant, and they align with each other. The network is then equivalent to a simple piecewise linear function. The second phase of training then reduces the error. Figures 4 and 5 in Appendix give an illustration of what is happening during the training process.

# 4 Main results

(This is a summary of 9, see there for details.)

## 4.1 Definitions

**Definition $\mathcal{S}_\sigma$**
Given $K \in \mathbb{N}$ and training data $(\vec{x}_k, y_k) \in \mathbb{R}^d \times \mathbb{R}$ for $k = 1, ..., K$, we can associate to each vector $\vec{w} \in \mathbb{R}^d$ the sequence of signs $F(\vec{w}) := (\sigma_1, ..., \sigma_K) \in \{-1, 0, +1\}$ given by $\sigma_k := \text{sign}(\langle \vec{w}, \vec{x}_k \rangle)$. We call the sequences of signs $\sigma$ that arise in this way the "sectors" defined by the training data, and we denote by $S_\sigma := F^{-1}(\sigma)$ the set of $\vec{w}$ that belong to this sector.

The sector of a weight vector $\vec{w}_i$ determines which training points $\vec{x}_k$ are "active" ($\sigma(k) = +1$) for that weight, which in turn determines how Gradient Descent will modify $\vec{w}_i$.

If all $\sigma_k \neq 0$, we call the sector $\sigma$ "open", since $S_\sigma$ is an open subset of $\mathbb{R}^d$. The other sectors are lower dimensional subsets of $\mathbb{R}^d$; together they give a partition of $\mathbb{R}^d$.

Here is an example with 8 open sectors (colored 2-dim cones) and 8 one-dimensional sectors (borders between cones):

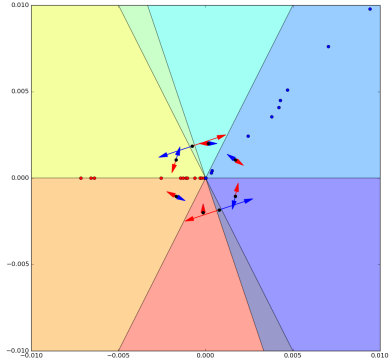

Independent of the dimension $d$ we can say that there are at most $3^K$ sectors. If the VC dimension of the input points $\vec{x}_k$ is $d'$, we can bound the number of sectors by $O(K^{d'})$, in particular (since $d' \leq d$) this is also $O(K^d)$.

For each sector $\sigma$ we now define a certain vector $\vec{D}_\sigma \in \mathbb{R}^d$ (essentially the initial velocity of a weight in this sector given by a "Gradient Descent differential equation on $S_\sigma$"):

**Definition $\vec{D}_\sigma$**
Given the loss function $\ell(\hat{y}, y)$, the training data $(\vec{x}_k, y_k)$, and a sector $\sigma$, define $V_\sigma \subset \mathbb{R}^d$ to be the vector space orthogonal to all $\vec{x}_k$ with $\sigma(k) = 0$, and denote by $\pi_\sigma : \mathbb{R}^d \to V_\sigma$ the orthogonal projection onto $V_\sigma$. (If $\sigma$ is an open sector, we have $V_\sigma = \mathbb{R}^d$ and $\pi_\sigma$ is the identity.) Then define

$$\vec{D}_\sigma := \sum_{k, \sigma(k) > 0} \frac{\partial}{\partial \hat{y}} \ell(0, y_k) \cdot \pi_\sigma(\vec{x}_k)$$

More generally, at each time point during training the weights define a function $f$, and we define

$$\vec{d}_\sigma := \sum_{k, \sigma(k) > 0} \frac{\partial}{\partial \hat{y}} \ell(f(\vec{x}_k), y_k) \cdot \pi_\sigma(\vec{x}_k)$$

**Definition** $G : \mathbb{R}^d \to \mathbb{R}$
We abbreviate $\vec{D}_{F(\vec{w})}$ by $\vec{D}_{\vec{w}}$. Then define: $G : \mathbb{R}^d \to \mathbb{R}$ by $G(\vec{w}) := \langle \vec{w}, \vec{D}_{\vec{w}} \rangle$.

On each sector, this is a linear function, and on all of $\mathbb{R}^d$ it is continuous since $\langle \vec{w}, \vec{x}_k \rangle = \langle \vec{w}, \pi_\sigma(\vec{x}_k) \rangle$ for $\vec{w} \in \mathcal{S}_\sigma$. We will be mostly interested in the values of $G$ on the unit sphere $\mathbb{S} \subset \mathbb{R}^d$. We will see that during the "initialization phase" the $\vec{u}_i := \vec{w}_i / |\vec{w}_i|$ move on $\mathbb{S}$ according to Gradient Ascent to increase $\text{sign}(w_i') \cdot G(\vec{u}_i)$.

**Definition** Extremal vectors
Let $\vec{w} \in \mathbb{R}^d$. Then we say "$\vec{w}$ is extremal" if $\vec{w} \neq 0$ and $\vec{w}/|\vec{w}|$ is a strict local maximum or minimum of $G : \mathbb{S} \to \mathbb{R}$. ("Strict" means the function is not constant in a neighborhood of the point.)

When a vector $\vec{w}$ is extremal, then $\vec{w} \in \vec{D}_{\vec{w}} \cdot \mathbb{R}$. Not all sectors have extremal vectors since often neither of $\pm \vec{D}_\sigma$ are in $S_\sigma$. But since the continuous function $G$ on the compact set $\mathbb{S}$ must have a minimum and a maximum, there exist sectors with extremal vectors unless $G$ is constant (which can only happen when all $\vec{D}_\sigma = 0$, which in turn only happens when all $f(\vec{x}_k) = y_k$).
If $0 \neq \vec{D}_\sigma \in \mathcal{S}_\sigma$, then we cannot have $-\vec{D}_\sigma \in \mathcal{S}_\sigma$ and vice versa, so for every extremal vector $\vec{D}_\sigma$ there is a unique sign $\tau_\sigma \in \{\pm 1\}$ with $\tau_\sigma \vec{D}_\sigma \in \mathcal{S}_\sigma$.

**Definition** $f_\mathcal{J}$
For each set $\mathcal{J}$ of sectors $\sigma$ that have extremal vectors $\tau_\sigma \vec{D}_\sigma \in \mathcal{S}_\sigma$ for $\tau_\sigma \in \{\pm 1\}$, we can look at a network with $|\mathcal{J}|$ neurons initialized with weights $\vec{w}_\sigma := \lambda \cdot \tau_\sigma \vec{D}_\sigma$ and $w_\sigma' := \lambda \cdot \tau_\sigma |\vec{D}_\sigma|$; after rescaling the initialization with $\lambda \to 0$ and training with "continuous Gradient Descent", this network describes a function $f_\mathcal{J}$.

For example, if the points $\vec{x}_1, ..., \vec{x}_K$ lie on a line, there are only $2K + 1$ sectors, and the functions $f_\mathcal{J}$ are certain piecewise linear functions with at most $2K + 1$ kinks.

In general, we only have $O(K^d)$ as an upper bound on the number of possible sectors; however, we observe experimentally (see Fig. 6) that the number of sectors with an extremal vector and the resulting number of kinks is much smaller; this number depends on the function that we are trying to fit, rather than only on the points $\vec{x}_k$.

We will also look at a modification, in which we "grow a network" one neuron at a time:

**Definition $\tilde{f}_\Sigma$**
For a sequence $\Sigma = (\sigma^{(1)}, \sigma^{(2)}, ..., \sigma^{(n)})$ of sectors, construct a ReLU network with one hidden layer in the following way:

- Start with an empty hidden layer,
- For $i = 1$ to $n$:
    - For the sector $\sigma := \sigma^{(i)}$ with extremal vector $\tau_\sigma \vec{d}_\sigma \in \mathcal{S}_\sigma$, $\tau_\sigma \in \{\pm 1\}$, add a hidden neuron with infinitesimal weights $\vec{w}_i := \lambda \cdot \tau_\sigma \vec{d}_\sigma$ and $w'_i := \lambda \cdot \tau_\sigma |\vec{d}_\sigma|$.
    - Run continuous Gradient Descent until convergence

The function $\tilde{f}_\Sigma$ is the limit for $\lambda \to 0$ of the resulting functions described by these networks.

Similar to the definition of $f_\mathcal{J}$ above, this definition assumes that the sectors used in each step have an extremal vector.
The difference to the definition of the $f_\mathcal{J}$ is that the $\vec{d}_\sigma$ change in each step, since the $f(\vec{x}_k)$ change by adding a neuron and optimizing. We then use these $\vec{d}_\sigma$ in the definition of $G$ and of the "extremal" vectors in each step.

## 4.2 Theorems

The main result is that the "simple" functions $f_\mathcal{J}$ already describe all possible functions we can get with Gradient Descent (from a network of arbitrary size) when both initialization and learning rate are small:

**Theorem 1**:
Given the training data $(\vec{x}_k, y_k)$ for $k = 1, 2, ..., K$, any ReLU network (except possibly a null set in the space of weights) with one hidden layer describes (after "continuous Gradient Descent") a function which in the limit $\lambda \to 0$ of small initialization converges to one of the finitely many functions $f_\mathcal{J}$.

The proof is given in the appendix 9. The key observation is that during the initial phase, the direction of the weights $\vec{w}_i$ converge to directions of extremal vectors $\tau_\sigma \vec{d}_\sigma$, and their magnitude $\log |\vec{w}_i|$ grows proportional to $|\vec{d}_\sigma|$, so they become relevant in the order of decreasing $|\vec{d}_\sigma|$. One further consequence of this proof is that each of these functions can also be obtained in the modified procedure:

**Theorem 2**:
Each function $f_{\mathcal{J}}$ is equal to some $\tilde{f}_{\Sigma}$ with $|\Sigma| \leq |\mathcal{J}|$, i.e. can also be obtained by "growing a network one neuron at a time" with at most $|\mathcal{J}|$ neurons.

In general, the sectors in $\Sigma$ differ from those in $\mathcal{J}$ (since during the optimization steps the weight vectors can move into other sectors).
The main reason for considering this modification is that the inclusion of Theorem 2 is strict: In general, not every $\tilde{f}_{\Sigma}$ can be obtained as a $f_{\mathcal{J}}$.

In fact, the mathematical idealization of "infinitesimal initialization" used in Theorem 1 is a bit too strict to reflect the practice of small initialization:
For example, if all $\vec{x}_k$ lie inside the unit ball (not counting the additional coordinate that we always set to 1), and all $y_k > 0$, then only the sector $\sigma = (1, 1, ..., 1)$ has an extremal vector. Infinitesimal initialization forces sectors without extremal vectors to become empty after the inital phase, so the function of the resulting network can be obtained as well by a network with only one neuron; this obviously would not be enough to approximate the values $y_k$.
But with more realistic small initialization, we have a shorter initialization phase and still can have some small weight vectors in every sector after the initialization phase. They can become important in the second phase when the loss is reduced and the dynamical system changes. This can be modeled in the above way as neurons that get added later with a new infinitesimal weight.

## 5 ReLU Networks and Linear Interpolations, Conjectures

The above result says that ReLU networks with one hidden layer, trained by Gradient Descent from infinitesimal initialization and learning rate, converge to "simple" functions which could also be given by a smaller number of neurons, depending only on the training data, but not on the size of the original network.

In this section we speculate about how those functions given by ReLU networks with infinitesimal initialization relate to linear interpolation.

Of course, any given ReLU network represents (by definition) a piecewise linear function, but this function depends on the network and can become arbitrarily complicated when the network is large.
The analogy is rather between linear interpolations and the ReLU network functions $f_{\mathcal{J}}$ (or $f_{\Sigma}$) from infinitesimal initializations - in both cases the training data and a finite number of choices define a function: For the case of linear interpolation the additional data are a triangulation of the points, for the networks it is the choice of sectors in $\mathcal{J}$ (or $\Sigma$).

For ReLU networks, we consider the two different models:

1. Infinitesimal initialization of a static network, functions $f_{\mathcal{J}}$:

Like linear interpolation, this procedure gives finitely many "simple functions" associated to training data $(\vec{x}_k, y_k)$. However, unlike linear interpolation it is too restricted to allow zero training error in general (as explained in 4.2).

2. Growing network with infinitesimal initialization, functions $\tilde{f}_\Sigma$:

    This is strictly more general than the previous model, since we are not restricted in the initialization of new neurons. Whenever such a network has not achieved zero training error, we can reduce the training error by adding a neuron in a sector that does not yet have $\vec{d}_\sigma = 0$.

    We conjecture that this process stops after $F_d(K)$ steps for some functions $F_d$. In fact, one may guess that $F_d(K)$ is of order $O(K)$, and that in practice we would see something between $K/d$ and $K$ steps. (This would be the case e.g. if we would fix each $\vec{w}_i$ once it is initialized to some $\tau_\sigma \vec{d}_\sigma$ and only optimize the scalars $w'_i$). If this is true, we get zero training error and have only finitely many choices for given training data, as for linear interpolation.

    Now the interesting question is whether this process approximates smooth functions $f : X \to \mathbb{R}$ when we sample randomly according to a "reasonable" probability distribution $\mathcal{D}_X$ on a "reasonable" bounded domain $X \subset \mathbb{R}^d$.

For the experiments, we consider two more variants:

2'. "Greedily" growing a network, special case of 2:
For linear interpolation there is (with probability 1, if the points are chosen according to some continuous probability distribution) one choice of triangulation that we can take as the "canonical" one: The Delaunay triangulation. Similarly, with probability 1 there is a "canonical" choice for the $\Sigma$ in 2: We always pick the sector $\sigma$ that has the largest $|\vec{d}_\sigma|$, corresponding to the biggest initial gain from adding this weight.
In practice, it is not feasible to go through all possible sectors to find the largest $|\vec{d}_\sigma|$, we rather take the largest $|\vec{d}_\sigma|$ from a random sample of sectors, see Algorithm 1.

3. Ensemble of growing networks, average of a set of $\tilde{f}_\Sigma$:
The above procedure gives the "minimal" networks / simplest functions that interpolate between the points, corresponding to very small initialization. In practice, we do not initialize quite so small and do not get "minimal" networks. One way to model this would be comparing it to an ensemble of our minimal models.
Qualitatively, this is not substantially different from the small growing networks, since for $K$ sample points it only averages at most $F_d(K)$ networks, so it would have the same theoretical properties.

But in experiments simulating "growing networks" we observed that they do generalize, but not as well as the networks initialized with "normal" initialization, whereas this ensemble model seems to give a generalization performance closer to the "normal" initialization.

# 6 Relation to Learning Theory

In Zhang et al. [2017] the authors report that on the one hand large neural networks generalize well on "real data", but on the other hand also give zero error approximations on randomly chosen $y_k$, which cannot generalize to other, unseen, randomly chosen labels $y$. This seems to suggest that neural networks may not fit the PAC learning model (appendix 10.1)

One can still save the PAC learning model by considering only neural networks that have some restriction on the weights, directly enforced by regularization, or indirectly by early stopping. This view is justified by the fact that these restrictions are used in practice.
However, this still leaves unexplained that even Gradient Descent without any regularization or early stopping can give results that generalize well to unseen test data.

Based on the conjecture / analogy made in the previous section, we here quickly discuss what can be said about generalization of linear interpolation and our Networks.

As in the case of linear interpolation (appendix 10.2) the above conjecture implies that for any given $K, d$, we have a bounded number of linear pieces, so we can construct a rapidly oscillating function that is approximated badly, so our ReLU networks cannot have the PAC learning guarantee.

So instead of looking at large ReLU networks as parametric models for arbitrary distributions and describing the function class that they could model, maybe we should look at them rather as non-parametric models like linear interpolation and see whether they have an approximation property for a smaller class of distributions (like those described by smooth functions with bounded second derivatives).

As an illustration, we would not expect this approximation property if the neural networks gave functions like this:

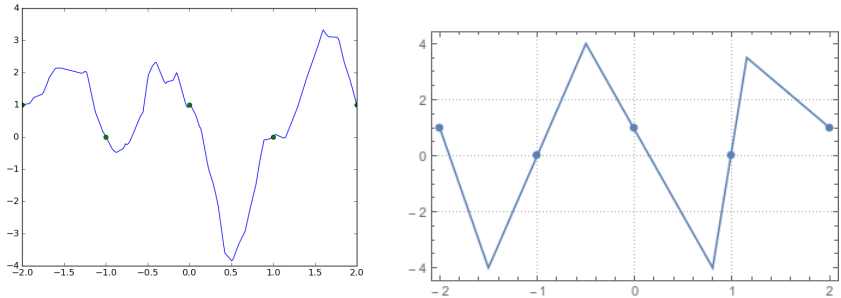

In these examples, we deviate significantly from the linear interpolation because we have "unnecessarily steep" parts between the points, which translates to large values of $\sum_i |w'_i| \cdot |\vec{w}_i|$. A (very speculative) reason to hope that the described process does not give such pictures is that each optimization step "tries to be economical with changes to $\sum_i |w'_i| \cdot |\vec{w}_i|$", and we start from an infinitesimal value. Whether this speculation turns out to be true remains to be seen.

If we can prove something like that, we are in the same situation as with linear interpolation: The general distribution-independent PAC learning property is violated, but a learned model can predict unseen labels in "reasonable" cases.

# 7 Experiments

The model described above is an idealized representation of what happens in practice. Both the weight initialization and the learning rate are small but not infinitesimal. In order to simulate the studied model experimentally, we use a greedy optimization approach where only $k < n$ neurons are trained at a time, with $k$ increasing until reaching a very small error on the training set. This simulates what gradient descent does when the weight initialization is infinitesimal. We report results on a 1d toy problem and on MNIST.

## 7.1 Replacement Network

When both the weight initialization and the learning rate go to zero, gradient descent leads to a two steps process (see subsection 4.2): (i) the weight vectors corresponding to each neuron is aligning to one $\vec{d}_i$, while the training error is staying approximately constant, (ii) the training error is decreasing. We simulate this process with the algorithm 1.

## 7.2 Setup

The experimental setup is as simple as possible, using basic versions of the usual algorithms and initialization schemes. We use a one hidden layer ReLU neural network to match the theoretical setup. The weights are initialized using truncated normals, with standard deviation of $magnitude/in\_dimension$, where

**Algorithm 1:** Greedy training algorithm simulating the idealized model studied in subsection 4.2.

    Initialize network with small random weights
    $k \leftarrow small\ number$
    $K \leftarrow large\ number$
    $mask \leftarrow \emptyset$
    $all\_neurons \leftarrow \{1, \ldots, n\}$
    **Function** `TrainNetwork`(*mask, steps*):
        **for** $i < steps$ **do**
            One Gradient Step for neurons in $mask$

    **Function** `ChooseNeuron`():
        **foreach** $i \in all\_neurons \setminus mask$ **do**
            $tmp\_mask \leftarrow mask \cup \{i\}$
            Save network in checkpoint `TrainNetwork`($tmp\_mask, k$)
            Restore network from checkpoint
        **return** *index with smallest loss*

    **while** *Training loss* $> \epsilon$ **do**
        $i \leftarrow$ `ChooseNeuron`()
        $mask \leftarrow mask \cup \{i\}$
        `TrainNetwork`($mask, K$)

$magnitude = 0.01$, except when it is the varying variable. We use $L2$ loss for the toy data, and cross entropy for MNIST. For the training procedure, we use Gradient Descent (*not* SGD) with a fixed step size, which is set to 0.1. There is no explicit regularization, and the network is trained until convergence on the training set. There is no use of a validation set, nor hyperparameter tuning [1].

### 7.3 Results

#### 7.3.1 Toy Data

To check that the algorithm behaves similarly to the idealized mathematical model, we run it on toy data in 1 dimension so that we can visualize the training.

**Initialization magnitude matters:** Section 3.2 and related figures show that the initialization has a large effect on the final function that the neural network produces after training. In particular, with large initialization, the function seems very arbitrary and what one would expect by the very large over-parametrization of the network. However, when the initialization is small, we get the very surprising effect that the network produces a simple piecewise linear often going through all the points[2].

---
[1] All those parameters have been set when running on a simple 1d example, and have been kept fixed since then. A few exploratory runs have shown that the exact values don't seem to matter for the phenomenon we explore, even though it could be interesting to explore that question further.

**Weight quantization happens:** Figures 4 and 5 show how the network function evolves over time. Those figures also include the position of the kinks of each neuron, and we see that they align to a small discrete set of positions, which in addition corresponds to the training points in that case. This is consistent with the mathematical model we propose and analyzed.

**Relationship to generalization:** To be more representative of the classifical learning setup, where the training points are sampled from an underlying distribution, and the question is how well the learnt function generalizes to new points from the distribution, we include figure 6. In this experiment, the underlying distribution is a simple piecewise linear function with 5 pieces in "w shape", and the training points are sampled from it (regression setup). The kinks of the neurons are also aligning to a discrete subset of the input space, and correspond to the inflection points of the underlying distribution. That means that the complexity of the learnt network does not depend on the number of training points, but on the properties of the underlying distribution.

### 7.3.2 MNIST

We also conduct experiments on the higher dimensional, yet simple, MNIST dataset. In particular, we want to estimate the gap of performance between usual Gradient Descent training and algorithm 1 which simulates our idealized mathematical model.

**Idealized model seems reasonable** Figure 1 shows the test accuracy after training on subsets of MNIST of different sizes, in function of the number of neurons. First, we observe that when the number of examples is large, both the normal training (*Full NN* on the figure) and the greedy training (*Greedy NN*) achieve very comparable and non trivial performances. When the number of examples is smaller, there is a clear gap between the two algorithms, which shows that our idealized model does not capture all the properties of GD, even though both still achieve good performance.

On figure 2, we observe that the number of active neurons reached by algorithm 1 is very small compared to the number of examples. That shows that our bound is very loose in this case, and that the complexity of the network is very small. Also, given that this number is an upper bound on the number of used neurons, independent of the starting size of the network, Fig. 1 also contains the performance of an ensemble of networks trained by the greedy algorithm. For fair comparison to the *Full NN* curve, the number of components in the ensemble is the number of neurons divided by the number of active neurons, so the total number of neurons in the ensemble is equal to the total number of neurons in the *Full NN*. Our hypothesis is that GD applied to a large NN partially creates an ensemble of smaller networks resembling the *Greedy NN* ones. Checking this hypothesis further requires more experimental and formal evidence.

---

[2]In all generality the optimization does not always find a zero loss solution, especially if the network is not highly overparametrized.

Finally, Fig. 3 shows the number of neurons needed to achieve 100% accuracy on the training set, as a function of the training set size. Again, very few neurons are needed to achieve that performance. Interestingly, both *Full NN* and *Greedy NN* are very similar on that metric.

# 8 Conclusions

We analyzed mathematically Gradient Descent on ReLU networks with one hidden layer and infinitesimal initialization and learning rate and showed that independent of the size of the network it converges to one of finitely many "simple" functions that depend only on the training data.

While these functions (like linear interpolations) do not satisfy a distribution independent generalization property, we ask whether they do satisfy an approximation property for a more restricted class of distributions, e.g. in the regression setting smooth functions with bounded second derivative. This would give some explanation of how ReLU networks of unbounded size can generalize to unseen inputs on "sensible" data while still being able to interpolate randomly choosen labels with zero training error.

# 9 Appendix A: Proof of theorems 1 and 2

## 9.1 Gradient descent for discrete time steps

Set
$e_k := \frac{\partial}{\partial \bar{y}} \ell(f(\vec{x}_k), y_k)$
$A_i := \{k \mid \langle \vec{w}_i, \vec{x}_k \rangle > 0\}$ ("active points" for neuron $i$)

Then
$$\frac{\partial \, L(f)}{\partial \ldots} = \sum_{k=1}^{K} -e_k \cdot \frac{\partial f(\vec{x}_k)}{\partial \ldots}$$

and with our $f$ from (1) we have

$$\frac{\partial f(\vec{x}_k)}{\partial w'_i} = \begin{cases} \langle \vec{w}_i, \vec{x}_k \rangle & \text{for } k \in A_i \\ 0 & \text{else} \end{cases} \qquad \nabla_{\vec{w}_i} f(\vec{x}_k) = \begin{cases} w'_i \cdot \vec{x} & \text{for } k \in A_i \\ 0 & \text{else} \end{cases}$$

Set $\vec{d}_A := \sum_{k \in A} e_k \cdot \vec{x}_k$ and $\vec{d}_i := \vec{d}_{A_i}$, then this gives the gradient of the loss

$$\frac{\partial L(f)}{\partial w'_i} = -\langle \vec{w}_i, \vec{d}_i \rangle \qquad \nabla_{\vec{w}_i} L(f) = -w'_i \cdot \vec{d}_i$$

Strictly speaking, $\nabla_{\vec{w}_i} L(f)$ is not always defined since $L(f)$ is not differentiable when for some $i, k$ we have $\langle \vec{w}_i, \vec{x}_k \rangle = 0$. However, we only arrive at such weights $\vec{w}_i$ with probability 0 if the initial values were chosen according to a continuous probability distribution. We will ignore all such events that only happen with probability 0.

However, when we later go over to "Continuous Gradient Descent" in 9.4, it will be important to correctly define the flow of $\vec{w}_i$ for *all* weight vectors, since then the probability that at some time we have $\langle \vec{w}_i, \vec{x}_k \rangle = 0$ is no longer zero.

This gives the Gradient Descent update rules:

**Lemma 1:**
For learning rate $\epsilon > 0$ one step in Gradient Descent is given by

$$w'_i \to w'_i + \epsilon \cdot \langle \vec{w}_i, \vec{d}_i \rangle \qquad \vec{w}_i \to \vec{w}_i + \epsilon \cdot w'_i \cdot \vec{d}_i \qquad (2)$$

## 9.2 Sectors

At any time point, the vectors $\vec{d}_i = \vec{d}_{A_i}$ appearing in the update rule are the same for all neurons $i$ which have the same set $A_i = \{ k \mid \langle \vec{w}_i, \vec{x}_k \rangle > 0\}$ of active training points, which in turn only depends on the direction of $\vec{w}_i$.

We can partition the set of all $\vec{w} \in \mathbb{R}^d$ into "sectors" depending on the signs of $\langle \vec{w}, \vec{x}_k \rangle$. A "sector" is then given by a sequence $\sigma = (\sigma_1, ..., \sigma_K) \in \{-1, 0, +1\}^K$ as
$$\mathcal{S}_\sigma := \{\vec{w} \in \mathbb{R}^d \mid \text{sign}(\langle \vec{w}, \vec{x}_k \rangle) = \sigma_k \text{ for } k = 1, 2, ..., K\}$$

We will assume in the following that the $\vec{x}_k$ are "in general position", i.e. no $d$ of these vectors are linearly dependent. (If the points $\vec{x}_k$ are randomly chosen with respect to some continuous probability distribution on $\mathbb{R}^d$, this is the case with probability 1.)

We have one sector given by $\sigma = (0, 0, ..., 0)$ that consists only of the origin. For the other sectors, the codimension of a $\mathcal{S}_\sigma$ is given by the number of "0" in $\sigma$. We call the sectors of codimension 0 also "open sectors" (since they are the ones that are open in the topology of $\mathbb{R}^d$).

In general, not all sequences $(\sigma_1, ..., \sigma_K) \in \{-1, 0, +1\}^K$ have a non-empty sector $\mathcal{S}_\sigma$, e.g. the codimension can be at most $d$, so if $\sigma \neq (0, 0, ..., 0)$ it can have at most $d - 1$ zeros.

For a subset $A \subseteq \{1, ..., K\}$ we can define the signs $\sigma_A$ by

$$\sigma_A(k) = \begin{cases} +1 & \text{for } k \in A \\ -1 & \text{else} \end{cases}$$

and we will denote the open sector $\mathcal{S}_{\sigma_A}$ also by $\mathcal{S}_A$.

We can visualize the $\vec{w}_i$ (red or blue dots, depending on the sign of $\vec{w}'_i$), $\vec{d}_A$ (blue arrows), and open sectors $S_A$ (colored parts of the plane) like this:

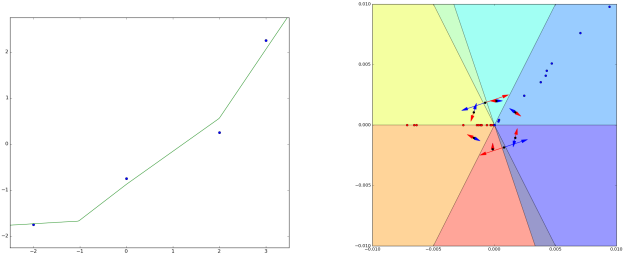

In each open sector, the weights represented by blue dots will move in the direction of the blue arrows (and red dots in the direction of the red arrows).

## 9.3 Vector fields on stratified spaces

There are different notions of stratified spaces; we don't need any general theory since we will only look at the above example of sectors in $V = \mathbb{R}^d$ and the induced structures on $W = \mathbb{R}^d \times \mathbb{R}$ and on the sphere $\mathbb{S} \subset \mathbb{R}^d$, which we describe here concretely.

A stratification of a space $X$ is a filtration by closed subsets

$$\emptyset = X_{-1} \subseteq X_0 \subseteq X_1 .... \subseteq X_n = X$$

For $V = \mathbb{R}^d$ we let $V_i$ be the union of all sectors of dimension $\leq i$. So $V_0$ consists of one point and $V_i$ for $0 < i < d$ is a union of finitely many sub-vectorspace of $V$ of dimension $i$. For $W = V \times \mathbb{R}$ we set $W_{i+1} = V_i \times \mathbb{R}$ and $W_0 = \emptyset$. For the unit sphere $\mathbb{S} \subset V$ we set $\mathbb{S}_i := V_{i+1} \cap \mathbb{S}$. For a sector $\mathcal{S}_\sigma$ of $V$, we call $\mathcal{S}_\sigma \times \mathbb{R}$ a "sector" of $W$ and $\mathcal{S}_\sigma \cap \mathbb{S}$ a "sector" of $\mathbb{S}$.

So in each case, the $X_i \backslash X_{i-1}$ are smooth $i$-dimensional manifolds, and the sectors are their connected components. Also, in each case $X$ itself is a smooth manifold, the stratification is only used to define "piecewise smooth" structures.

We will call a function $f$ from a stratified space $X$ to $\mathbb{R}$ "piecewise differentiable" if it is differentiable on each $\bar{\mathcal{S}}_\sigma$ in all directions inside $\bar{\mathcal{S}}_\sigma$.

Our main example will be the function $g : V \to \mathbb{R}$ given by $\vec{w} \mapsto \langle \vec{w}, \vec{d}_A \rangle$ on the closure of the open sectors $\bar{\mathcal{S}}_A$. Obviously, it is differentiable on the closure of each sector, so it only remains to show that the definitions coincide on $\bar{\mathcal{S}}_A$ and $\bar{\mathcal{S}}_{A'}$ when $\vec{w} \in \bar{\mathcal{S}}_A \cap \bar{\mathcal{S}}_{A'}$.

Let $I$ be the set of indices $k$ that are only in one of the two sets $A, A'$. Then $\vec{w} \in \bar{\mathcal{S}}_A \cap \bar{\mathcal{S}}_{A'}$ implies on the one hand that $\langle \vec{w}, \vec{x}_k \rangle = 0$ for $k \in I$, and on the other hand $\vec{d}_A$ and $\vec{d}_{A'}$ differ only by a linear combination of those $\vec{x}_k$, so indeed $\langle \vec{w}, \vec{d}_A \rangle - \langle \vec{w}, \vec{d}_{A'} \rangle = \langle \vec{w}, \vec{d}_A - \vec{d}_{A'} \rangle = 0$.

A stratified differential form $\omega$ on $X$ is given by differential forms $\omega_\sigma$ on the closure of each sector $\bar{\mathcal{S}}_\sigma$ such that for $\vec{x} \in \bar{\mathcal{S}}_\sigma \subset \bar{\mathcal{S}}_{\sigma'}$ the restriction of $\omega_{\sigma'}$ to the tangential space at $\vec{x}$ in $\bar{\mathcal{S}}_\sigma$ is equal to $\omega_\sigma$. By our definitions, $df$ is a stratified differential form for any piecewise differentiable function $f$ on $X$.

The dual notion are stratified vector fields:
A system of continuous functions $v_\sigma : \bar{\mathcal{S}}_\sigma \to T\bar{\mathcal{S}}_\sigma$ is called a stratified vector field if for all $\bar{\mathcal{S}}_\sigma \subset \bar{\mathcal{S}}_{\sigma'}$ we have $\pi(v_{\sigma'}|_{\bar{\mathcal{S}}_\sigma}) = v_\sigma$ for the orthogonal projection $\pi : T\bar{\mathcal{S}}_{\sigma'} \to T\bar{\mathcal{S}}_\sigma$.

On $V, W, \mathbb{S}$ we have a standard Riemannian metric (for $V, W$ the Euclidean metric, for $\mathbb{S} \subset V$ the restriction of the Euclidean metric to the tangential bundle of $\mathbb{S}$), and thus also a canonical isomorphism $\sharp : T^*X \to TX$. In this notation the gradient of a function is given by $\nabla f = (df)^\sharp$.

The definition of stratified vector fields is such that the isomorphism $\sharp$ gives a bijection between stratified differential forms and stratified vector fields. In particular, for a piecewise differentiable function $f : X \to \mathbb{R}$ the gradient $\nabla f$ is a stratified vector field.

For our function $g : V \to \mathbb{R}$ the gradient is given by $\nabla g(\vec{w}) = \vec{d}_A$ on open sectors $\mathcal{S}_A$. We can describe the corresponding $v_\sigma$ concretely: For a sector $\sigma$ let $V_\sigma$ be the orthogonal component of all $\vec{x}_k$ with $\sigma(k) = 0$, this is the tangent space $T\mathcal{S}_\sigma$, and let $\pi_\sigma : V \to V_\sigma$ be the orthogonal projection. Then for any open sector $\mathcal{S}_A$ and any sector $\sigma$ with $\mathcal{S}_\sigma \subseteq \bar{\mathcal{S}}_A$ the vector field $v_\sigma$ is the constant

$$\vec{d}_\sigma := \pi_\sigma(\vec{d}_A)$$

Since it is the gradient of a piecewise differentiable function, this definition of $\vec{d}_\sigma$ must be independent of the choice of $A$.

## 9.4 Gradient Descent differential equation

We will now look at how the $\vec{w}_i$ change over time and derive the differential equation corresponding to (2) for $\epsilon \to 0$.

This is obvious as long as $\vec{w}_i$ is inside an open sector $\mathcal{S}_A$:

$$\frac{\partial}{\partial t} w'_i = \langle \vec{w}_i, \vec{d}_A \rangle \qquad \frac{\partial}{\partial t} \vec{w}_i = w'_i \cdot \vec{d}_A$$

The solution of this differential equation will match the solution of the discrete equation (2) for $\epsilon \to 0$.

However, we also have to determine the behavior for the case that $\vec{w}_i$ is at the boundary between two (or more) open sectors. In the discrete case (2) $\vec{w}_i$ can change frequently between different sectors:

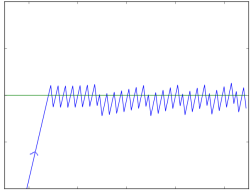

In this case $\vec{w}$ is trapped on (or near) the boundary and can only move in direction of the boundary. However, since the $\vec{d}_\sigma$ form a stratified vector field on $V$, the movement in direction of the boundary is the same on both sides of the boundary, so we can describe this movement by a differential equation on the boundary.

**Lemma 2:**
The "Continuous Gradient Descent" (i.e. Gradient Descent for learning rate $\epsilon \to 0$) is given by the stratified vector field $\vec{d}_\sigma$ on $V = \mathbb{R}^d$ that defines a "stratified differential equation" on $W = \mathbb{R}^d \times \mathbb{R}$

$$\frac{\partial}{\partial t} w'_i = \langle \vec{w}_i, \vec{d}_\sigma \rangle \qquad \frac{\partial}{\partial t} \vec{w}_i = w'_i \cdot \vec{d}_\sigma \qquad (3)$$

A solution of this stratified differential equation is given by a piecewise differentiable curve $(\vec{w}_i(t), w'_i(t))$ (each piece in one of the sectors), which satisfies the differential equation (9.6) when $\vec{w}_i(t)$ is in sector $\sigma$. When this sector is not open, we additionaly require that all neighboring sectors $\sigma'$ with $\dim(\sigma') = \dim(\sigma) + 1$ have a $\vec{d}_{\sigma'}$ that points towards $\sigma$ (i.e. "$\vec{w}_i(t)$ cannot leave $\sigma$ because it would immediately come back"). This specifies uniquely a solution to our differential equation when a starting point is given, unless at some time point $t$ two neighboring sectors switch the component of $\vec{d}$ orthogonal to the boundary from pointing towards their boundary to pointing away from the boundary. That these two quantities switch sign at the same time usually has probability 0 (if the coefficients of the system are chosen randomly according to a continuous distribution), so in "almost all" cases the solution is uniquely defined by its starting point.

So we have on each sector $\mathcal{S}_\sigma$ a differential equation of the same type, and we defined an overall "solution of the stratified differential equation" (9.6) such that it matches also the movement of a $\vec{w}_i$ "trapped near $\mathcal{S}_\sigma$" for $\epsilon \to 0$.

### 9.5 Invariant quadratic form, polar coordinates

We now show how we can reduce our stratified differential equation on $\mathbb{R}^d \times \mathbb{R}$ to a stratified differential equation on the $(d-1)$-dimensional unit sphere.

Our stratified differential equation (9.6) is of a special form that has $w_i'^2 - |\vec{w}_i|^2$ as conserved quantities:

$$\begin{aligned}
\frac{\partial}{\partial t}\Big(w_i'^2 - \langle \vec{w}_i, \vec{w}_i \rangle\Big) &= 2\Big(w_i' \cdot \frac{\partial}{\partial t} w_i' - \langle \frac{\partial}{\partial t} \vec{w}_i, \vec{w}_i \rangle\Big) \\
&= 2\Big(- w_i' \langle \vec{w}_i, \vec{d}_i \rangle + \langle w_i' \cdot \vec{d}_i, \vec{w}_i \rangle\Big) = 0
\end{aligned}$$

Since we are interested in the case $\lambda \to 0$, we initialize with very small $w_i, \vec{w}_i$, and so these differences will be small. In the following we will assume that it actually is zero, which corresponds to setting $w_i' = \pm |\vec{w}_i|$ at initialization time. (This is not really needed, but simplifies the following arguments somewhat.)

With probability one all $w_i', \vec{w}_i \neq 0$ at initialization, and our differential equation conserves this property, so we will assume that as well in the following. If $w_i'$ is never 0, this also means that the signs $s_i := \text{sign}(w_i')$ remain constant. Together with the invariance property above, that means that $w_i'$ is already determined if we know $\vec{w}_i$ and this constant $s_i \in \{\pm 1\}$, so we only need to consider the $\vec{w}_i$ in our differential equation.

Since $\vec{w}_i \neq 0$ we can introduce "polar coordinates" in our weight space: Let $\mathbb{S}$ be the $d-1$ dimensional unit sphere in $\mathbb{R}^d$, then we can write each weight vector as $\vec{w}_i = e^{r_i} \cdot \vec{u}_i$ with $r_i := \log |\vec{w}_i| \in \mathbb{R}$ and $\vec{u}_i := \vec{w}_i \cdot e^{-r_i} \in \mathbb{S}$.

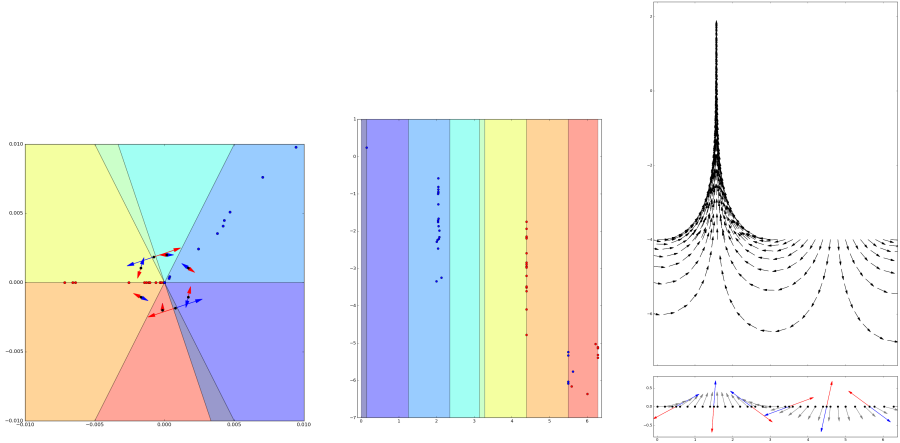

We can express the stratified differential equation for $\vec{w}_i$ in these coordinates:

$$\frac{\partial}{\partial t}(e^{r_i} \cdot \vec{u}_i) = \frac{\partial}{\partial t} r_i \cdot e^{r_i} \vec{u}_i + e^{r_i} \cdot \frac{\partial}{\partial t} \vec{u}_i = w'_i \cdot \vec{d}_i$$

Since $\frac{\partial}{\partial t} \vec{u}_i$ must be tangential to the unit sphere and hence orthogonal to $\vec{u}_i$, we get by taking the scalar product with $\vec{u}_i$:

**Lemma 3a:**
In polar coordinates the evolution of the weights $\vec{w}_i$ under Continuous Gradient Descent is given by

$$\frac{\partial}{\partial t} r_i = s_i \cdot \langle \vec{d}_i, \vec{u}_i \rangle \qquad \frac{\partial}{\partial t} \vec{u}_i = s_i \cdot \left( \vec{d}_i - \langle \vec{d}_i, \vec{u}_i \rangle \vec{u}_i \right) \qquad (4)$$

In particular, the direction in which points move does not depend on the (log) radius $r$, and it depends on the angle $\vec{u}$ in a very simple way.

When we restrict our piecewise differentiable function $g : V \to \mathbb{R}$ to $\mathbb{S}$ and note that the projection of $\vec{d}_i = \nabla g(\vec{u}_i)$ to the tangent space to $\mathbb{S}$ at $\vec{u}_i$ is $\vec{d}_i - \langle \vec{d}_i, \vec{u}_i \rangle \vec{u}_i$, we can interpret the second part of (4) as follows:

**Lemma 3b:**
The directions $\vec{u}_i$ of the weights in the Continuous Gradient Descent satisfy the stratified differential equation on $\mathbb{S}$

$$\frac{\partial}{\partial t} \vec{u}_i = s_i \cdot \nabla_\mathbb{S} \, g(\vec{u}_i) \qquad (5)$$

where $\nabla_\mathbb{S}$ means the gradient on the Riemannian manifold $\mathbb{S}$.

This means the $\vec{u}_i$ are simply "following either Gradient Ascent or Gradient Descent for $g(\vec{u}_i)$ on $\mathbb{S}$". Of course, this is a slight abuse of language - over time the $\vec{d}_i$ and hence also the function $g$ change. But we will see that this is almost literally true in the initial phase, which we will analyze in the next section.

## 9.6 Initial phase for "$\lambda = 0$"

If we initialize the weights at $\lambda w'_i, \lambda \vec{w}_i$ with $\lambda$ very small, the resulting function values $f(\vec{x}_k)$ will be $O(\lambda^2)$, so the terms $e_k := \frac{\partial}{\partial \tilde{y}} \ell(f(\vec{x}_k), y_k)$ are in good approximation given by $\frac{\partial}{\partial \tilde{y}} \ell(0, y_k)$, and they do not change until some weight increases to a significant size.

To model this first phase for $\lambda \to 0$, we can set $E_k := \frac{\partial}{\partial \tilde{y}} \ell(0, y_k)$ and

$$\begin{aligned}
\vec{D}_A &= \sum_{k \in A} E_k \cdot \vec{x}_k \\
\vec{D}_\sigma &:= \pi_\sigma(\vec{D}_A) \quad \text{for } \mathcal{S}_\sigma \subset \bar{\mathcal{S}}_A \\
G(\vec{U}) &:= \langle \vec{U}, \vec{D}_{\vec{U}} \rangle \quad \text{with } \vec{D}_{\vec{U}} := \vec{D}_{F(\vec{U})}
\end{aligned}$$

so $\vec{D}_\sigma$ and $G$ do not depend on the time and we have a stratified differential equation with constant coefficients

$$\frac{\partial}{\partial t} \vec{U}_i = s_i \cdot \nabla_\mathbb{S} G(\vec{U}_i) = s_i \cdot \left( \vec{D}_i - \langle \vec{D}_i, \vec{U}_i \rangle \cdot \vec{U}_i \right) \tag{6}$$

We will first analyze the evolution of randomly initialized weight directions $\vec{U}_i$ in this model.

There are two differential equations of this type - one for $s_i = 1$, and one for $s_i = -1$, otherwise the differential equations for the $U_i$ only differ in their initial value. So we will assume $s \in \{\pm 1\}$ fixed and drop the index $i$.

We expect these directions $\vec{U}$ to accumulate at some fixed point of this motion, i.e. at zeros of $\nabla_\mathbb{S} G$; however, we need some quantitative estimate how fast they move to prove they really "arrive there". For this we compute the change of the quantity $s \cdot G(\vec{U})$ over time from (6):

$$\begin{aligned}
\frac{\partial}{\partial t} s \cdot G(\vec{U}) &= s \cdot \langle \frac{\partial}{\partial t} \vec{U}, \vec{D}_{\vec{U}} \rangle \\
&= \langle \vec{D}_{\vec{U}} - \langle \vec{D}_{\vec{U}}, \vec{U} \rangle \cdot \vec{U}, \ \vec{D}_{\vec{U}} \rangle \\
&= |\vec{D}_{\vec{U}}|^2 - \langle \vec{U}, \vec{D}_{\vec{U}} \rangle^2 \\
&= |\vec{D}_{\vec{U}}|^2 \cdot \sin^2(\vec{U}, \vec{D}_{\vec{U}}) \tag{7}
\end{aligned}$$

This also describes the fixed points:
This expression is 0 iff either $\vec{D}_{\vec{U}} = 0$ or $\vec{U} \in \mathbb{R} \cdot \vec{D}_{\vec{U}}$. Both can happen:

On the one hand, a weight could start on $\mathbb{R} \cdot \vec{D}_\sigma$ for some sector $\sigma$; however, this has probability zero, so we will assume this is not the case.

On the other hand, all sectors without active training points, i.e. the ones with all signs $\sigma_k \in \{0, -1\}$ will have $\vec{D}_\sigma = 0$ (and of course this remains true during all the time even for the non-constant differential equation). Some weights may start in a sector $\mathcal{S}_\sigma$ with $\vec{D}_\sigma = 0$, and some may get into one of these sectors during the initial phase. If that happens they don't change any more, remain very small, and in the end they don't contribute to the $f$, so we can disregard them.

With probability one the other sectors have initially $\vec{D}_\sigma \neq 0$, and we will assume that in the following. Let $D_{min}$ be the minimum of the $|\vec{D}_\sigma|$ with $\vec{D}_\sigma \neq 0$.

For any given $\delta > 0$ we can use (7) to bound the time it will take until all weights have reached $\angle(\vec{U}, \vec{D}_{\vec{U}}) < \delta$ from the $\vec{D}_{\vec{U}}$ of their sector: Let $\Delta G := \max_{\vec{u} \in \mathbb{S}} G(\vec{u}) - \min_{\vec{u} \in \mathbb{S}} G(\vec{u})$. Then we have to wait at most

$$\Delta t(\delta) := \Delta G \cdot (D_{min} \cdot \sin(\delta))^{-2}$$

until all $\vec{U}$ are or have been within an angle $\delta$ to their $\vec{D}_{\vec{U}}$. Even after they reached a $\vec{D}_\sigma$ within that angle, they still can change the sector while increasing $G(\vec{U})$, if the new sector $\sigma'$ has a larger $\vec{D}_{\sigma'}$. However, since the number of sectors is finite, they can do so only a bounded number of times, and each time after $\Delta t(\delta)$ they are or have been within $\delta$ of the new $\vec{D}_\sigma$. So at most after $\#sectors \cdot \Delta t(\delta)$ all $\vec{U}$ are within angle $\delta$ of their $\vec{D}_{\vec{U}}$ and cannot leave this neighborhood of $\vec{D}_{\vec{U}}/|\vec{D}_{\vec{U}}|$ any more. If $\delta$ is smaller than the minimal angle between two different non-zero $\vec{D}_\sigma$, then this also means the sector of $\vec{U}$ cannot change any more.

We can summarize the result as follows:

**Lemma 4a:**

Given training data $\vec{x}_k, y_k$; assume the "infinitesimal initialization model", in particular (6) is valid with a constant $G$.

Then for every $\delta > 0$ there is a time $T_\delta$ such that all weight directions $\vec{U} \in \mathbb{S}$ (except for a set with probability 0) after time $t \geq T_\delta$ arrived in a sector $\mathcal{S}_\sigma$ that they do not leave any more. These sectors either have

- $\vec{D}_\sigma = 0$, and $\vec{U}$ remains constant for $t \geq T_\delta$, or

- there is an extremal vector $\pm \vec{D}_\sigma \in \mathcal{S}_\sigma$ with $\angle(\vec{U}, \pm \vec{D}_\sigma) < \delta$, and $\vec{U}$ stays in this neighborhood of $\pm \vec{D}_\sigma/|\vec{D}_\sigma|$ for all $t \geq T_\delta$.

Once $\vec{U}$ is in a fixed sector, $\vec{D} := \vec{D}_{\vec{U}}$ is a constant and we can solve the original equations

$$\frac{\partial}{\partial t} w' = \langle \vec{w}, \vec{D} \rangle \qquad \frac{\partial}{\partial t} \vec{w} = w' \cdot \vec{D}$$

explicitly: Set $a := w'(t_0)$ and $b := \langle \vec{w}(t_0), \frac{\vec{D}}{|\vec{D}|} \rangle$, then

$$w'(t+t_0) = a \cdot \cosh(t \cdot |\vec{D}|) + b \cdot \sinh(t \cdot |\vec{D}|)$$
$$\vec{w}(t+t_0) = \vec{w}(t_0) + \left( a \cdot \sinh(t \cdot |\vec{D}|) + b \cdot \cosh(t \cdot |\vec{D}|) - b \right) \cdot \frac{\vec{D}}{|\vec{D}|}$$

We now look at how this changes when we multiply the weight initializations with $\lambda \to 0$. To get a limiting function we will at the same time adjust the time scale by subtracting $\log(\lambda)/|\vec{D}|$ and use the invariants $a' := a/\lambda$, $b' := b/\lambda$. Since for $t \to \infty$ we have $e^{-t} \cdot \cosh(t) \to 1$, $e^{-t} \cdot \sinh(t) \to 1$, we get for $\lambda \to 0$

$$w'\left(t + t_0 - \frac{\log(\lambda)}{|\vec{D}|}\right) \to (a' + b') \cdot e^{t \cdot |\vec{D}|}$$
$$\vec{w}\left(t + t_0 - \frac{\log(\lambda)}{|\vec{D}|}\right) \to (a' + b') \cdot e^{t \cdot |\vec{D}|} \cdot \frac{\vec{D}}{|\vec{D}|}$$

In particular this gives

**Lemma 4b:**
For each $i = 1,..,n$ there are constants $c_i$ and sectors $\sigma_i$ such that for every time $t$ after initialization with $\vec{w}_i = \lambda \cdot \vec{v}_i$, $w'_i = s_i \cdot |\vec{w}_i|$ at $t=0$ and running the idealized Continuous Gradient Descent using the constant $\vec{D}_\sigma$, we have in the limit $\lambda \to 0$

$$\vec{w}_i\left(t - \frac{\log(\lambda)}{|\vec{D}_{\sigma_i}|}\right) \to e^{t \cdot |\vec{D}_{\sigma_i}|} \cdot c_i \cdot \vec{D}_{\sigma_i}$$

## 9.7 Initial phase for "$\lambda \to 0$"

The next step is to compare this idealized "$\lambda = 0$" version using $E_k = \frac{\partial}{\partial \hat{y}} \ell(0, y_k)$ with the real "$\lambda \to 0$" dynamics using $e_k = \frac{\partial}{\partial \hat{y}} \ell(f(\vec{x}_k), y_k)$. Here we need to make some assumptions on the loss function. In the two cases we are mostly interested in, we have:

- Regression ($y \in \mathbb{R}$) with L2 loss:
  $\frac{\partial}{\partial \hat{y}} \ell(\hat{y}, y) = \hat{y} - y$

- Classification ($y \in \{\pm 1\}$) with cross entropy loss:
  $\frac{\partial}{\partial \hat{y}} \ell(\hat{y}, y) = -y \cdot \sigma(-y\hat{y})$ with $\sigma(t) := 1/(1+e^{-t})$ the logistic sigmoid.

Both functions are Lipschitz continuous as functions of $\hat{y}$ with a constant $K$ (independent of $y$: $K = 1$ for regression, $K = 1/4$ for classification). So for simplicity we will assume such a property for our loss functions (although the results could still be proven for much weaker assumptions if necessary).

The other ingredient is a bound on $f(\vec{x})$, which we get directly from the definition: Since we have $|w'_i| = |\vec{w}_i|$, the definition of $f$ gives us

$$|f(\vec{x}_k)| \leq |\vec{x}_k| \cdot \sum_i |\vec{w}_i|^2$$

Together with the assumption on the loss function this gives for the difference of the real $\vec{d}_A$ to the $\vec{D}_A$ in the idealized situation for any open sector $\mathcal{S}_A$:

$$|\vec{d}_A - \vec{D}_A| \leq |A| \cdot K \cdot |\vec{x}_k|^2 \cdot \sum |\vec{w}_i|^2 = c_1 \cdot \sum |\vec{w}_i|^2$$

with some constant $c_1 > 0$ which depends on the inputs $\vec{x}_k$ and the loss function $\ell$ (but not on the $\vec{w}_i$). The same is true for the lower-dimensional sectors $\sigma$, since their $\vec{d}_\sigma$ and $\vec{D}_\sigma$ are just orthogonal projections of some $\vec{d}_A$ and $\vec{D}_A$. We can also use that to bound $\vec{d}_\sigma$:

$$|\vec{d}_\sigma(t)| \leq |\vec{D}_\sigma| + c_1 \cdot \sum_i |\vec{w}_i(t)|^2 \leq |\vec{D}_{max}| + c_1 \cdot \sum_i e^{2 \cdot r_i(t)}$$

We can use that in turn to bound $r_i(t)$: Let $r(t)$ be the maximum of the $r_i(t)$. Then

$$\frac{\partial}{\partial t} r(t) \leq |\vec{D}_{max}| + c_1 \cdot n \cdot e^{2 \cdot r(t)} \leq c_2 \cdot (1 + e^{2 \cdot r(t)})$$

So if we start with $r(0) \ll 0$, $r(t)$ cannot grow much faster than $c_2 \cdot t$. We can make that precise with the solutions $\tilde{r}(t)$ of

$$\frac{\partial}{\partial t} \tilde{r}(t) = c_2 \cdot (1 + e^{2 \cdot \tilde{r}(t)})$$

which are given by

$$\tilde{r}_a(t) = c_2 \cdot (t - a) - \frac{1}{2} \log\left(1 - e^{2c_2(t-a)}\right)$$

These functions are defined for $t < a$, they are just the translates of $r_0$, which is monotonically increasing and has $\lim_{t \to -\infty} r_0(t) = -\infty$ and $\lim_{t \to 0} r_0(t) = \infty$, so for any given initial value $r(0)$ we have a unique $a$ such that $\tilde{r}_a(0) = r(0)$, and then we have

$$r_i(t) \leq \tilde{r}_a(t) \Rightarrow |\vec{d}_\sigma - \vec{D}_\sigma| \leq c_1 \cdot n \cdot e^{2 \cdot \tilde{r}_a(t)} \qquad \text{for } t \in [0, a[$$

By multiplying the initialization weights with $\lambda \to 0$, we get $r(0) \to -\infty$ and thus $a \to -\infty$, from which we can also deduce that after time $T_\delta$ we have for all sectors $\sigma$

$$|\vec{d}_\sigma - \vec{D}_\sigma| \leq c_1 \cdot n \cdot e^{2 \cdot \tilde{r}_a(T_\delta)} \to 0 \qquad \text{for } \lambda \to 0$$

Since the $\vec{d}_\sigma$ define the stratified differential equation for $\vec{w}_i$, for $\lambda \to 0$ also the solutions $\vec{w}_i$ of the "finite $\lambda$" differential equation converge to the solutions of the "$\lambda = 0$" differential equations.

So for $\lambda \to 0$ we get the same result as for the idealized scenario "$\lambda = 0$":

**Lemma 5:**
Given training data $\vec{x}_k, y_k$; assume the derivative of the loss function $\frac{\partial}{\partial \hat{y}} \ell(\hat{y}, y)$ is Lipschitz continuous (as is the case for L2 loss and cross entropy loss).
Choose vectors $\vec{v}_i \in \mathbb{R}^d$ and signs $s_i \in \{\pm 1\}$. For a $\lambda > 0$ initialize the weights by $\vec{w}_i := \lambda \cdot \vec{v}_i$, $w'_i := s_i \cdot \lambda \cdot |\vec{v}_i|$ and run Continuous Gradient Descent.
Then for $\lambda$ small enough and every $\delta > 0$ there is a time $T_\delta$ such that (with probability 1) all weight directions $\vec{u}_i = \vec{w}_i/|\vec{w}_i|$ arrived in a sector $\mathcal{S}_\sigma$ with either

- $\vec{d}_\sigma = 0$, and $\vec{u}_i$ remains constant for $t \geq T_\delta$, or

- there is an extremal vector $\pm \vec{d}_\sigma \in \mathcal{S}_\sigma$ and an extremal vector $\pm \vec{D}_\sigma \in \mathcal{S}_\sigma$ of the constant system such that $\angle(\vec{u}_i, \pm \vec{d}_\sigma) < \delta$ and $|\vec{D}_\sigma - \vec{d}_\sigma| < \delta$

## 9.8 Construction of a network defining the limit function

The main part of the proof done, it remains to assemble the results into the theorems 1 and 2 of 4.2. To write down the remaining arguments formally would require some more work, for now we will be content with a reduced level of formality (a.k.a. "physicist's proof") for the remaining arguments.

From Lemma 5 we know that for every neuron $i$ there is a sector $S_{\sigma_i}$ such that in the initialization phase $\vec{w}_i$ will end up in sector $S_{\sigma_i}$; and for any $\delta > 0$ small enough, after running the differential equation for a time $T_\delta$, all $\vec{w}_i$ will be in sector $S_{\sigma_i}$ and in direction of $\tau_\sigma \vec{d}_{\sigma_i}$, $\tau_\sigma \in \{\pm 1\}$ up to an angle of $\delta$.

**Def.:** Let $\mathcal{J}$ be the set of all these sectors $\sigma_i$.

Furthermore, the magnitudes $r_i$ grow proportional to $t \cdot |\vec{D}_{\sigma_i}|$, so they become significant in the order of decreasing $|\vec{D}_{\sigma_i}|$. With probability 1, the numbers $|\vec{D}_\sigma|$ are all different for different sectors $\sigma$, we will assume that in the following. (This is only for notational convenience, we could also in the definition of $\tilde{f}_\Sigma$ use a sequence of groups of sectors in $\Sigma$ to avoid this assumptions.)
By reordering the $\vec{w}_i$ if necessary, we can assume that they are ordered such that
$$|\vec{D}_{\sigma_1}| \geq |\vec{D}_{\sigma_2}| \geq .... \geq |\vec{D}_{\sigma_n}|$$
By making $\lambda$ smaller by some factor $\lambda'$ and increasing $t$ by subtracting $\log(\lambda')/|\vec{D}_{\sigma_1}|$ to compensate, we can make the difference between different $t|\vec{D}_\sigma|$ as large as needed, so that only one new group of weights corresponding to one sector $\sigma$ enters the optimizition at a time while the next group of sectors is still too small to matter.

**Def.:** Let $\Sigma$ be the sequence of these sectors as they become significant.

Now we can define the networks for $f_\mathcal{J}$ and $\tilde{f}_\Sigma$ as in 4.1. The claim of theorem 1 is that in the limit $\lambda \to 0$ the original network describes the function $f_\mathcal{J}$, and theorem 2 follows from the claim that $\tilde{f}_\Sigma = f_\mathcal{J}$.

These functions $f_\mathcal{J}$ and $\tilde{f}_\Sigma$ are modelled after the original network for infinitesimal initialization, Lemma 4 describes the initial phase (which also applies to the initial phase of each step in the case of $\tilde{f}_\Sigma$), Lemma 5 describes that this also applies to finite, but small $\lambda$, what remains to show is that after the initial phase produced results close to those of the models for $f_\mathcal{J}$, $\tilde{f}_\Sigma$, that the resulting functions remain close also during the optimization.

## 9.9 Comparison to finite $\lambda$

After the initial phase we follow the differential equation (9.6); a standard theorem for ordinary differential equations says that the solution depends continuously on the initial conditions (see e.g. Walter [1998], p.141). Assuming this carries over to stratified differential equations, this means also that we can adjust $\lambda$ (by a factor) and the time scale (by a summand) such that for any time point the solutions for finite $\lambda$ converge to the idealized solution, which would give us Theorem 1 and 2.

For the dependence of the solution to a stratified differential equation from the initial conditions we have to look at the flow across boundaries.
When the neighborhood of a solution moves "into" a lower dimensional boundary sector, it collapses in the direction orthogonal to the boundary, so this still depends continuously on the initial conditions.
When a solution moves "out" of a lower dimensional boundary sector, it follows an ordinary differential equation in the new sector. (This assumes that we have a solution to the stratified differential equation, i.e. we exclude the case that two neighboring sectors $\sigma, \sigma'$ switch the $\vec{d}_\sigma$ and $\vec{d}_{\sigma'}$ from pointing towards the boundary to pointing away from the boundary at the same time. Since $d_\sigma$ and $d'_\sigma$ involve different $\vec{x}_k$, this means that this will happen with probability 0 if the input data are randomly generated from a continuous distribution).
When the solution moves "across" a lower dimensional boundary and the component away from the boundary increases in the direction it is going, the neighborhood is getting longer by the factor given by the ratio of the two components. But this is a finite factor, so the solution still depends continuously on the initial conditions.

So the continuous dependence on the initial conditions remains true for stratified differential equations, which means that even after the optimization step(s) the function of the original network converges for $\lambda \to 0$ to $f_\mathcal{J}$ and $\tilde{f}_\Sigma$, which gives Theorems 1 and 2.

# 10 Appendix B: Learning Theory and Linear Interpolations

## 10.1 PAC learning

We are here interested in a learning algorithm that takes as an input a sequence $z_k = (\vec{x}_k, y_k)$ of samples, which we assume are randomly sampled from some probability distribution $\mathcal{D}$ on $Z = X \times Y$ and outputs a function $h$ ("hypothesis") in some function space $\mathcal{H}$. The quality of this output is judged by some loss function $\ell : \mathcal{H} \times Z \to \mathbb{R}$ from which we can define the "true" error/risk as $L_\mathcal{D}(h) := \mathbb{E}_{z \sim \mathcal{D}}[\ell(h, z)]$.

In the case of regression without noise, the distribution $\mathcal{D}$ on $Z$ would be given by a distribution $\mathcal{D}_X$ on $X$ and a (deterministic) function $f : X \to Y = \mathbb{R}$.

A standard approach to showing that the learning algorithm "works" is to prove a generalization property that is true for all distributions $\mathcal{D}$, e.g. "Agnostic PAC Learnability"(Shalev-Shwartz and Ben-David [2014], p.25):
$\exists m_\mathcal{H} : (0,1)^2 \to \mathbb{N}$, s.t. $\forall \epsilon, \delta \in (0,1), \forall m \geq m_\mathcal{H}(\epsilon, \delta), \forall \mathcal{D}$ :
using $m$ i.i.d. samples $\sim \mathcal{D}$, the algorithm returns $h$, s.t. with prob. $\geq 1 - \delta$:

$$L_\mathcal{D}(h) \leq \min_{h' \in \mathcal{H}} L_\mathcal{D}(h') + \epsilon \qquad (8)$$

(i.e. the algorithm probably gives a hypothesis in $\mathcal{H}$ which is almost as good as the best in $\mathcal{H}$.)

However, not all algorithms that "work" have this property. The simplest example would be the nearest neighbor algorithm that assigns to each $\vec{x}$ the value $y_k$ of the nearest sample point $\vec{x}_k$: For functions $f : [0,1] \to \mathbb{R}$ that are Lipschitz continuous with Lipschitz constant $\leq C$, increasing the number of samples $m \to \infty$ gives approximations that converge against the true function. However, for each $m$ we can give a function (e.g. a rapidly oscillating function) such that any piecewise constant function with $\leq m$ pieces will be a bad approximation (i.e. have an approximation error bounded away from zero). Similar statements are true for linear interpolation.

## 10.2 Linear Interpolation

Assume $X = [0,1]$ and the true function $f : [0,1] \to \mathbb{R}$ is differentiable and the derivative is Lipschitz continuous with Lipschitz constant bounded by $C$, then it is easy to see that for $0 \leq a \leq x \leq b \leq 1$:

$$\left| f(x) - (x-a) \cdot \frac{f(b) - f(a)}{b - a} \right| \leq 2C(b-a)^2$$

So with increasing number of samples $m \to \infty$ the linear interpolation converges to $f$ since the length of the longest sub-interval goes to 0 (in fact, almost surely like $\log(m)/m$, but we don't need that).

However, we need to restrict the function class encoded in the distribution $\mathcal{D}$ and use the constant $C$ from that $\mathcal{D}$, so this is weaker than the PAC learnability. In fact, given any $m$, we can choose a rapidly oscillating function $f$ which is Lipschitz with large constant, e.g. the function $f(x) = \sin(x \cdot 4m\pi)$, then any piecewise constant function with $\leq m$ pieces has an approximation error $\geq 1/48$ (see 10.3). On the other hand, since this $f$ is Lipschitz, it can be approximated arbitrarily well by functions in $\mathcal{H}$, thus $\min_{h' \in \mathcal{H}} L_{\mathcal{D}}(h') = 0$, violating (8)

In dimensions $d > 1$, we no longer interpolate between 2 points, but between up to $d+1$ points. There is no longer one unique linear interpolation which would be determined by the choice of sample points, but as additional data we need to give a triangulation. For a given set of points there are only finitely many choices for this triangulation, but in general each of them will give a different linear interpolation function.
For points in general position, there is one particular "Delaunay" triangulation that one can choose to define a unique interpolation.

If we assume $X = [0,1]^d$ and the true function $f : [0,1] \to \mathbb{R}$ is differentiable and the derivative is Lipschitz continuous with Lipschitz constant $\leq C$, we can choose a fine enough grid of points in $X$ and get linear interpolations that are arbitrary close to $f$ with the same arguments as in the one-dimensional case, and also by essentially the same argument we see that for a fixed $m$ we always find functions in the same function class (but high $C$) that cannot be approximated well.

It is more difficult to prove an approximation guarantee for more general bounded $X$ and randomly chosen sample points (according to some distribution $\mathcal{D}_X$) with the same argument - one would need some bound on the edge lengths of the triangulation.
There are such results, e.g. the results in Arkin et al. [2015] should be sufficient for $\mathcal{D}$ the uniform distribution on a unit ball in $\mathbb{R}^d$, and the Delaunay triangulation.

## 10.3 Rapidly oscillating functions

We have

$$\sin(x) \geq 0.5 \text{ for } x \in \left[\frac{\pi}{3}, 2 \cdot \frac{\pi}{3}\right], \quad \text{and} \quad \sin(x) \leq -0.5 \text{ for } x \in \left[4 \cdot \frac{\pi}{3}, 5 \cdot \frac{\pi}{3}\right]$$

Therefore, in any interval $[a,b]$, we have subintervals of total length $\geq \lfloor b-a \rfloor / 6$ in which $\sin(x \cdot 2\pi) \geq 0.5$, and subintervals of total length $\geq \lfloor b-a \rfloor / 6$ in which $\sin(x \cdot 2\pi) \leq -0.5$.
For a function $f : [a,b] \to \mathbb{R}$ that does not change the sign, we therefore must have
$$\int_a^b \left(f(x) - \sin(x \cdot 2\pi)\right)^2 dx \geq \frac{1}{4} \cdot \frac{\lfloor b-a \rfloor}{6} = \frac{\lfloor b-a \rfloor}{24}$$

and for $C > 1$

$$\int_a^b \left(f(x) - \sin(C \cdot x \cdot 2\pi)\right)^2 dx \geq \frac{\lfloor C \cdot (b-a) \rfloor}{24 \cdot C} \geq \frac{b-a}{24} - \frac{1}{24 \cdot C}$$

If we have a function $f : [0, 1] \to \mathbb{R}$ that changes signs at most $k$ times, we can apply this inequality to the (at most $k + 1$) parts with equal sign and get

$$\int_0^1 \left(f(x) - \sin(C \cdot x \cdot 2\pi)\right)^2 dx \geq \frac{1}{24} \cdot \left(1 - \frac{k+1}{24 \cdot C}\right)$$

In particular, for $C \geq 2k + 2$ we get

$$\int_0^1 \left(f(x) - \sin(C \cdot x \cdot 2\pi)\right)^2 dx \geq \frac{1}{48}$$

# 11  Appendix C: Figures

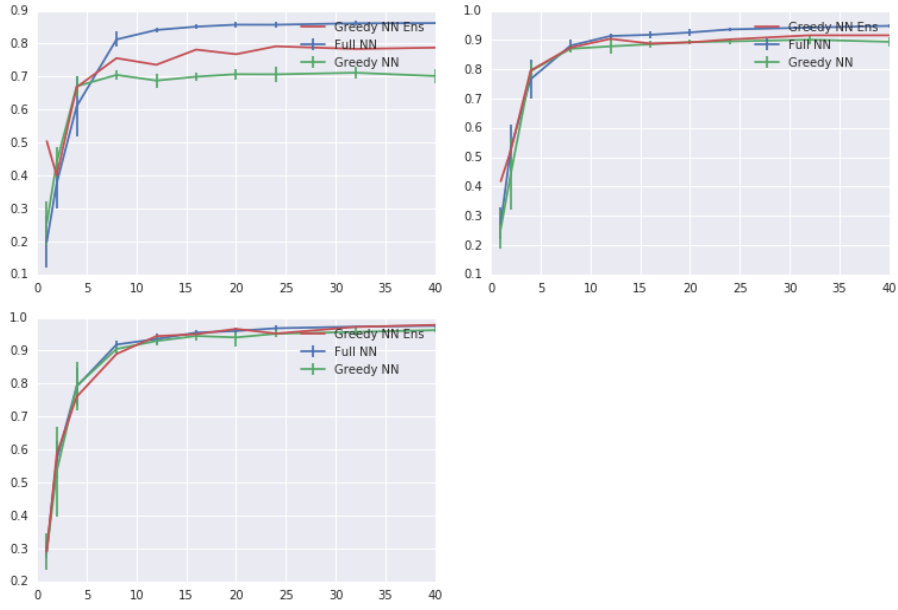

Figure 1: Test accuracies on MNIST in function of the number of neurons, using respectively 1000, 10000 and 60000 training examples. *Full NN* corresponds to a normal training using all neurons from the start. *Greedy NN* corresponds to the greedy training algorithm 1, simulating the idealized mathematical model. *Greedy NN Ens* is an ensemble of $k = \frac{num\_neurons}{num\_active}$ *Greedy NN* independent runs.

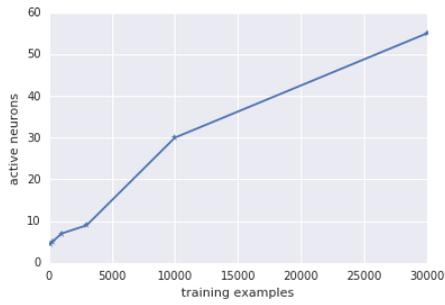

Figure 2: Number of active neurons in function of the number of MNIST training examples, when following algorithm 1, simulating the idealized mathematical model.

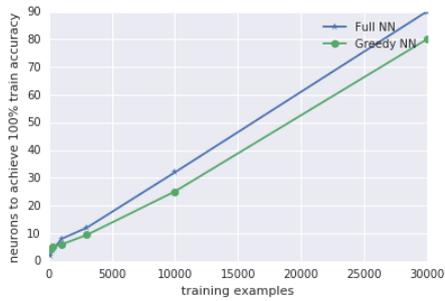

Figure 3: Number of neurons needed to achieve 100% training accuracy in function of the number of MNIST training example. *Full NN* corresponds to a normal neuron network and *Greedy NN* corresponds to the greedy training algorithm 1, simulating the idealized mathematical model.

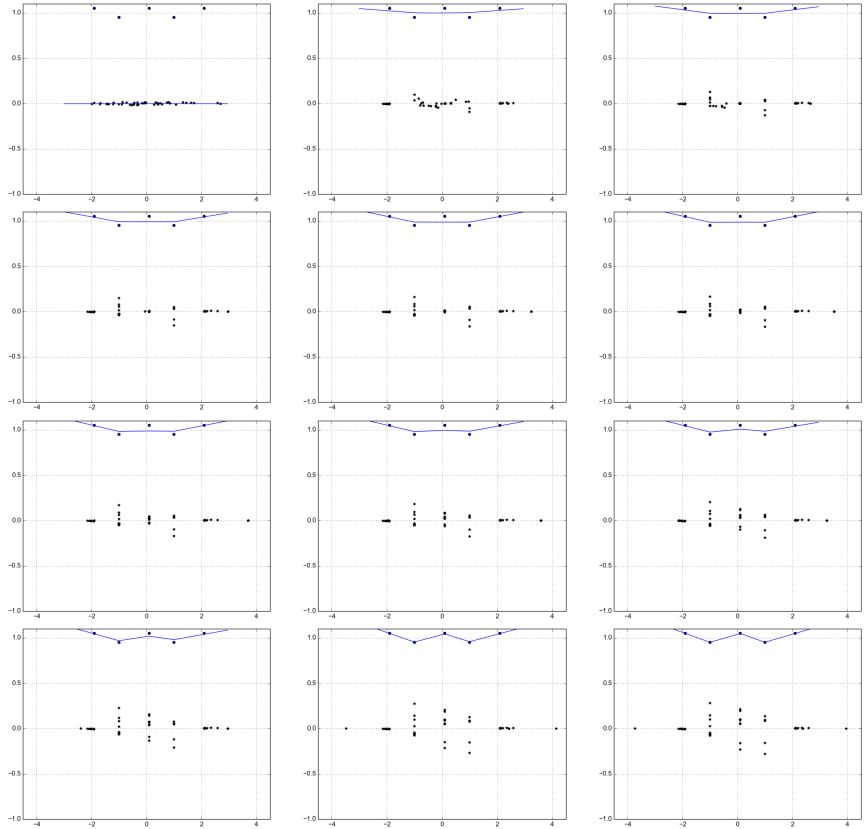

Figure 4: Frame based animation of a representative training run of a ReLU fully connected neural network using GD in 1d. The 5 blue points are the training points. The blue line is the output function of the network, which has one hidden layer and 50 units. The black stars represent the input weights to the hidden layer: the X-axis corresponds to the position of the ReLU kinks, i.e. when the ReLU input is 0; the Y-axis corresponds to the weight value. We clearly see an alignment of the kinks to a discrete set of positions, except for kinks corresponding to very small weights that stay small. In addition, the non zero kinks are aligned to the training points in that example. The final function is a simple piecewise linear function going through the points, with the minimal number of pieces.

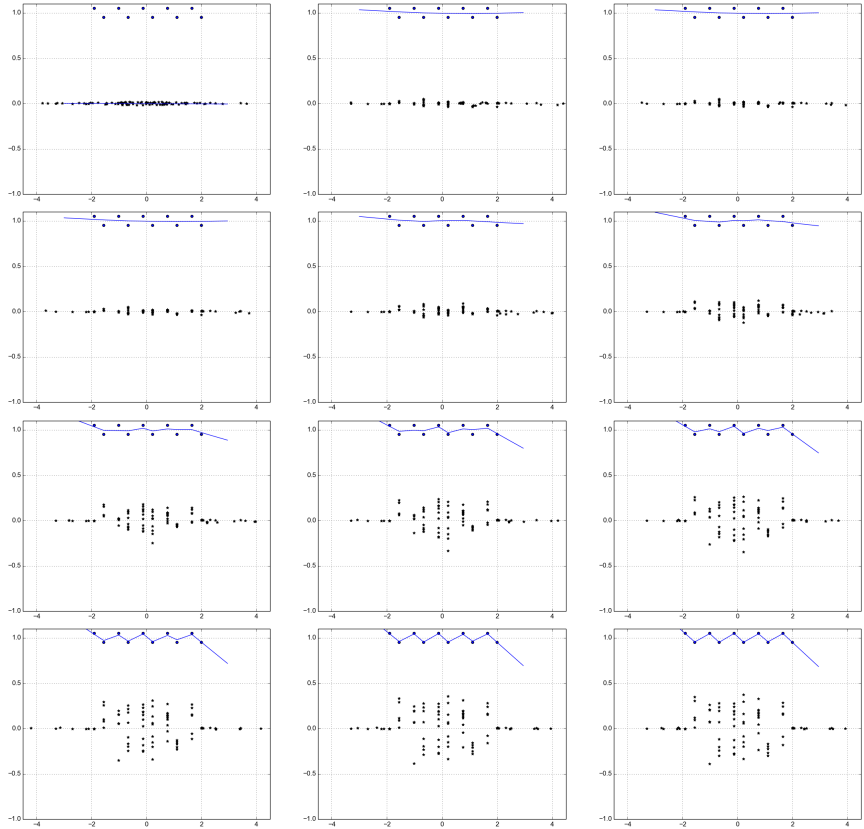

Figure 5: Frame based animation of a representative training run of a ReLU fully connected neural network using GD in 1d. The 10 blue points are the training points. The blue line is the output function of the network, which has one hidden layer and 100 units. The black stars represent the input weights to the hidden layer: the X-axis corresponds to the position of the ReLU kinks, i.e. when the ReLU input is 0; the Y-axis corresponds to the weight value. We clearly see an alignment of the kinks to a discrete set of positions, except for kinks corresponding to very small weights that stay small. In addition, the non zero kinks are aligned to the training points in that example. The final function is a simple piecewise linear function going through the points, with the minimal number of pieces.

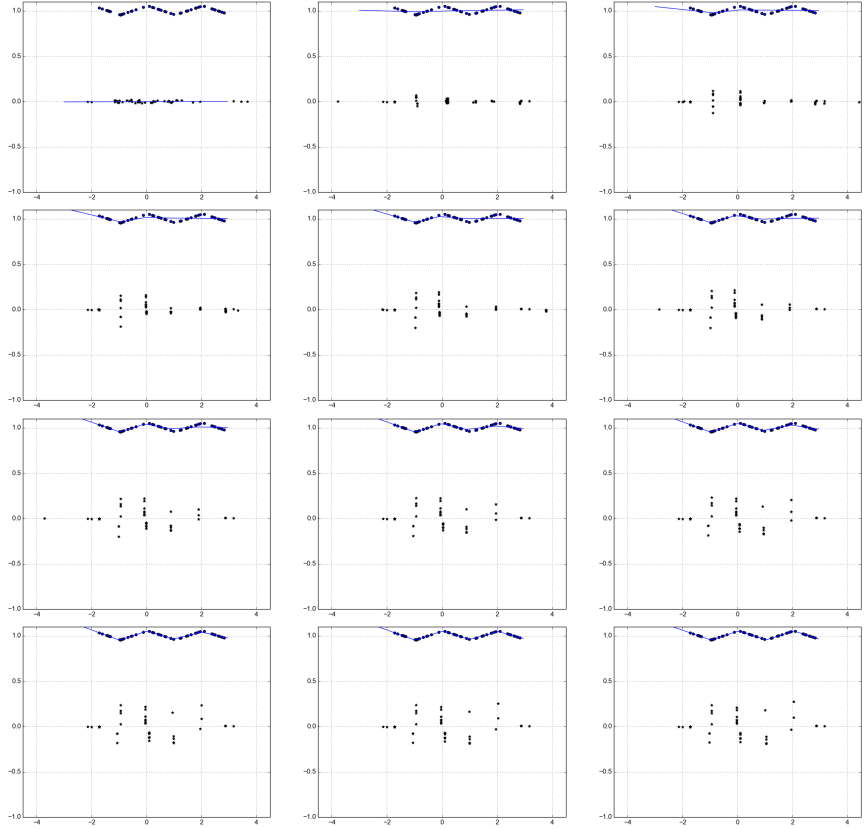

Figure 6: Frame based animation of a representative training run of a ReLU fully connected neural network using GD in 1d. The blue points are the training points. They are distributed uniformly on a simple piecewise linear function with 5 pieces. The blue line is the output function of the network, which has one hidden layer and 50 units. The black stars represent the input weights to the hidden layer: the X-axis corresponds to the position of the ReLU kinks, i.e. when the ReLU input is 0; the Y-axis corresponds to the weight value. We clearly see an alignment of the kinks to a discrete set of positions, except for kinks corresponding to very small weights that stay small. In addition, the non zero kinks are aligned to the turning points. The final function is a simple piecewise linear function going through the points, with almost the minimal number of pieces. In this case, the function generated by the network would generalize very well, as it almost exactly recovered the underlying data distribution.